%% file: oov_object_detection.tex
\documentclass[a4paper, 10pt, conference]{ieeeconf}      
\usepackage{FG2019}

\FGfinalcopy 

\IEEEoverridecommandlockouts                              
\def\FGPaperID{20} 

\title{\LARGE \bf
Extended Gaze Following: \\Detecting Objects  in Videos Beyond the Camera Field of View
}

\author{\parbox{16cm}{\centering
    {\large Benoit Mass\'e$^{1}$, St\'ephane Lathuili\`ere$^{1,2}$, Pablo Mesejo$^{1,3}$ and Radu Horaud$^{1}$}\\
    {\normalsize
      $^1$ Inria \& Univ. Grenoble Alpes, France,
      $^2$ University of Trento, Italy,
      $^3$ University of Granada, Spain
  }}
  \thanks{This work is supported by ERC Advanced Grant VHIA \#340113.}
}

\usepackage{gensymb}
\usepackage{amssymb}
\usepackage{amsmath}
\usepackage{math} 
\usepackage{newtxmath}
\DeclareMathAlphabet{\mathpzc}{T1}{pzc}{m}{it}

\usepackage{multirow}
\usepackage{booktabs}

\usepackage{graphicx}
\usepackage[subrefformat=parens,labelformat=parens,lofdepth,lotdepth]{subfig}

\definecolor{colorbenoit}{RGB}{225, 51, 153}
\definecolor{colorbenoit2}{RGB}{153, 51, 245}
\makeatletter
\def\benoit{\@ifnextchar[{\@benoitopt}{\@benoitnoopt}}
\def\@benoitopt[#1]#2{\textcolor{colorbenoit}{[BENOIT: #1]} \textcolor{colorbenoit2}{[#2]}}
\def\@benoitnoopt#1{\textcolor{colorbenoit}{[BENOIT]} \textcolor{colorbenoit2}{[#1]}}
\makeatother

\def\eg{\emph{e.g}. } 
\def\ie{\emph{i.e}. }

\newcommand{\vern}{\emph{Vernissage} }

\newcommand{\synt}{\emph{synthetic} }

\graphicspath{{images/}}

\begin{document}


\ifFGfinal
\thispagestyle{empty}
\pagestyle{empty}
\else
\author{Anonymous FG 2019 submission\\ Paper ID \FGPaperID \\}
\pagestyle{plain}
\fi
\maketitle

\begin{abstract}


  In this paper we address the problems of detecting objects of interest in a video and of estimating their locations, solely from the gaze directions of people present in the video. Objects can be indistinctly located inside or outside the camera field of view. We refer to this problem as \emph{extended gaze following}. The contributions of the paper are the followings. First, we propose a novel spatial representation of the gaze directions adopting a top-view perspective. Second, we develop several convolutional encoder/decoder networks to predict object locations and compare them with heuristics and with classical learning-based approaches. Third, in order to train the proposed models, we generate a very large number of synthetic scenarios employing a probabilistic formulation. Finally, our methodology is empirically validated using a publicly available dataset.

\end{abstract}


\input{sec_introduction}

\input{sec_related}

\input{sec_model}

\input{sec_data}

\input{sec_experiments}

\input{sec_conclusion}





\bibliographystyle{unsrt}
\bibliography{biblio}
\section*{Supplementary Material}
\subsection{Ablation study: $T$}

\begin{figure}[!h]
\centering
\includegraphics[width=0.95\linewidth]{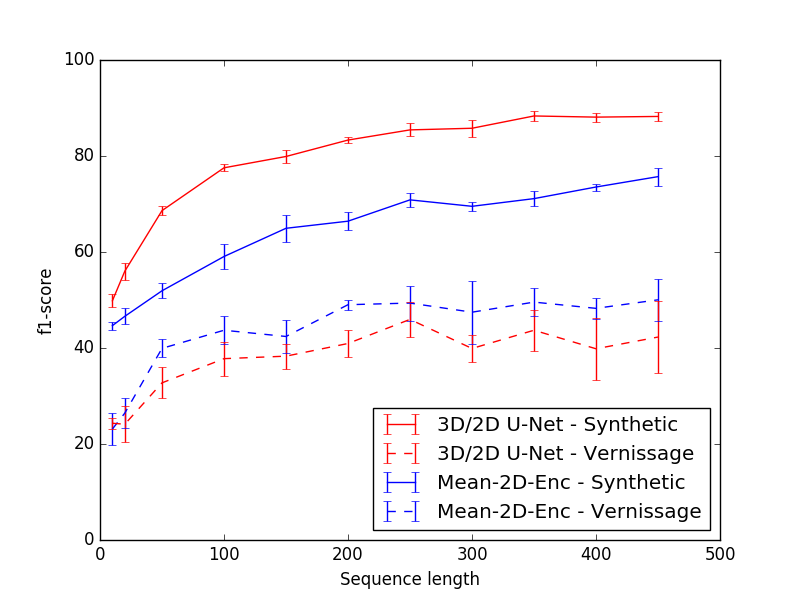}
\caption[Influence of the sequence length $T$ on the performances]{
  Performance obtained on the \synt and \vern datasets with RGB data. We measure the \emph{f1-score} with different values of sequence length $T$.
  \label{fig:Tvar}
}
\end{figure}

We report experiments to measure the impact in performance of the sequence length $T$ in Fig.~\ref{fig:Tvar}. Precisely, we selected \emph{Mean-2D-Enc} (as best model on \vern) and \emph{3D/2D U-Net} (as best model on \synt) and compute the \emph{f1-score} evolution for these two networks varying $T$ from 10 to 450. Both networks behave similarly to the results reported before: \emph{3D/2D U-Net} is consistently better on \synt data than \emph{Mean-2D-Enc}, and consistently worse on the \vern dataset. We observe that the performances of both networks tend to increase with the sequence length on \synt data, though quite slowly for $T>150$. However, when the networks are transfered to be used on the \vern dataset, the \emph{f1-score} stops increasing past $T=200$ or $250$. Moreover, the variances are sometimes quite higher, which could indicate a more unstable training process. This validates the choice of $T=200$ for our experiments.

\newpage

\subsection{Other synthetic examples}

\begin{figure}[!h]
\centering     
\subfloat[][\emph{Object heat-map}]{\label{fig:o_synt_1}
    \includegraphics[height=0.27\linewidth]{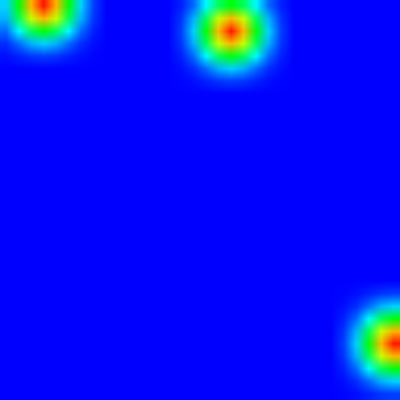}
} \hspace{0.2cm}
\subfloat[][\emph{Gaze heat-map}]{\label{fig:hm_synt_1}
    \includegraphics[height=0.27\linewidth]{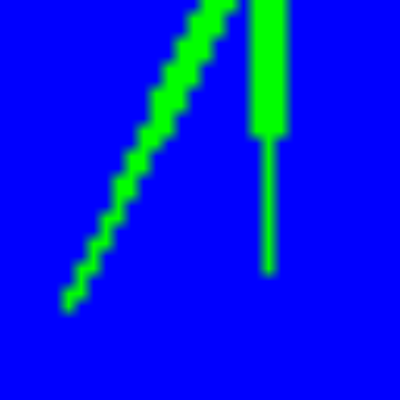}
} \hspace{0.2cm}
\subfloat[][Mean \emph{gaze heat-map}]{\label{fig:m_hm_synt_1}
    \includegraphics[height=0.27\linewidth]{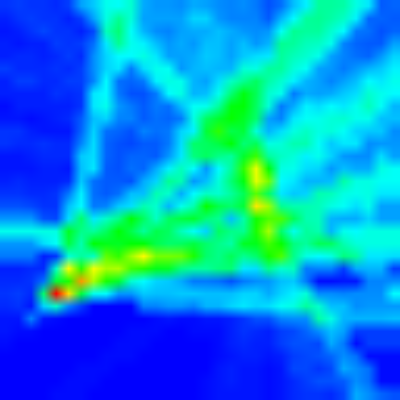}
}
\caption[Heat-maps from an example synthetic scenario ($N=2$ and $M=3$)]{Heat-maps from a synthetic scenario generated randomly, with $2$ people ($N=2$) and $3$ objects ($M=3$).~\subref{fig:o_synt_1}: the ground truth \emph{Object heat-map} $\Omegavect$ used for training or evaluation.~\subref{fig:hm_synt_1}: a \emph{Gaze heat-map} randomly chosen among the sequence.~\subref{fig:m_hm_synt_1}: the mean \emph{gaze heat-map} over the sequence.
}
\label{fig:illustration_synt_1}

\subfloat[][\emph{Object head-map}]{\label{fig:o_synt_2}
    \includegraphics[height=0.27\linewidth]{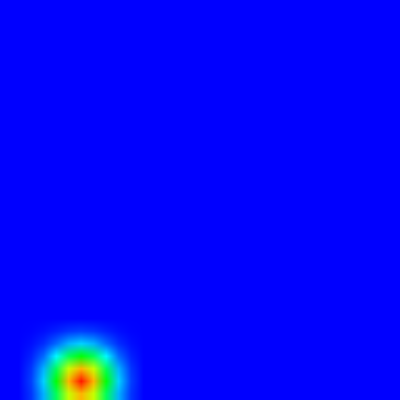}
} \hspace{0.2cm}
\subfloat[][\emph{Gaze heat-map}]{\label{fig:hm_synt_2}
    \includegraphics[height=0.27\linewidth]{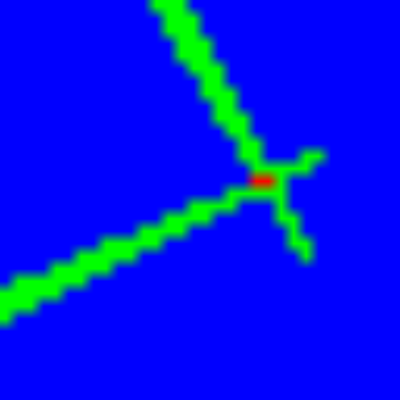}
} \hspace{0.2cm}
\subfloat[][Mean \emph{gaze heat-map}]{\label{fig:m_hm_synt_2}
    \includegraphics[height=0.27\linewidth]{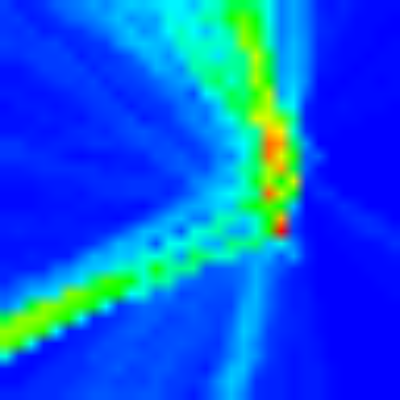}
}
\caption[Heat-maps from an example synthetic scenario ($N=2$ and $M=1$)]{Heat-maps from a synthetic scenario generated randomly, similar to Fig.~\ref{fig:illustration_synt_1}, but with a different setup: $2$ people ($N=2$) and $1$ object ($M=1$).
}
\label{fig:illustration_synt_2}

\subfloat[][\emph{Object head-map}]{\label{fig:o_synt_3}
    \includegraphics[height=0.27\linewidth]{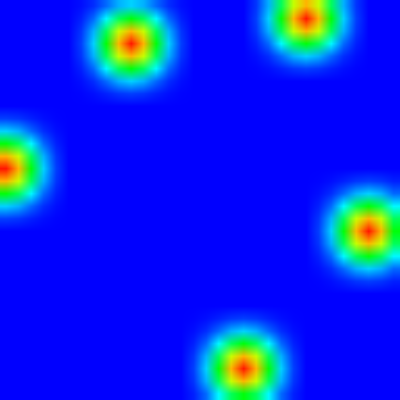}
} \hspace{0.2cm}
\subfloat[][\emph{Gaze heat-map}]{\label{fig:hm_synt_3}
    \includegraphics[height=0.27\linewidth]{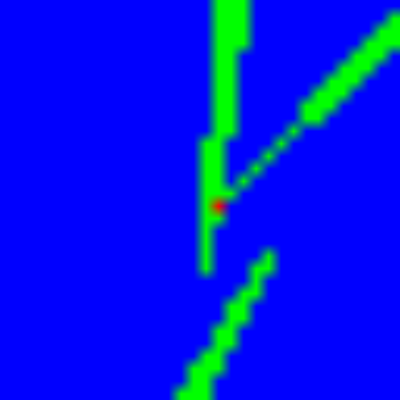}
} \hspace{0.2cm}
\subfloat[][Mean \emph{gaze heat-map}]{\label{fig:m_hm_synt_3}
    \includegraphics[height=0.27\linewidth]{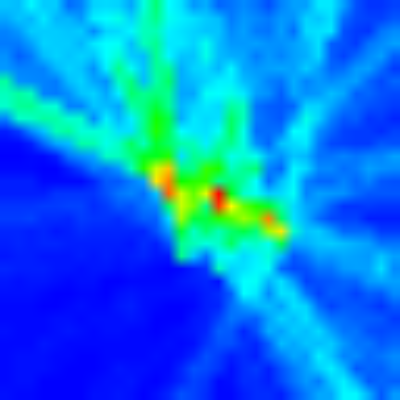}
}
\caption[Heat-maps from an example synthetic scenario ($N=3$ and $M=5$)]{Heat-maps from a synthetic scenario generated randomly, similar to Fig.~\ref{fig:illustration_synt_1}, but with a different setup: $3$ people ($N=3$) and $5$ objects ($M=5$).
}
\label{fig:illustration_synt_3}
\end{figure}

Example of generated scenarios in Fig.~\ref{fig:illustration_synt_1}-\ref{fig:illustration_synt_2}-\ref{fig:illustration_synt_3}. Fig.~\ref{fig:illustration_synt_1} is the generated scenario used in the paper.

\begin{figure}
\centering     
\subfloat[][$\hat{\Omegavect}$ - \emph{Mean-2D-Enc}]{\label{fig:omega_2D_synt}
    \includegraphics[height=0.30\linewidth]{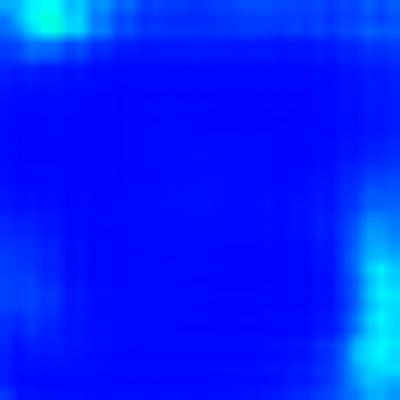}
}
\subfloat[][$\hat{\Omegavect}$ - \emph{3D/2D U-Net}]{\label{fig:omega_3D_synt}
    \includegraphics[height=0.30\linewidth]{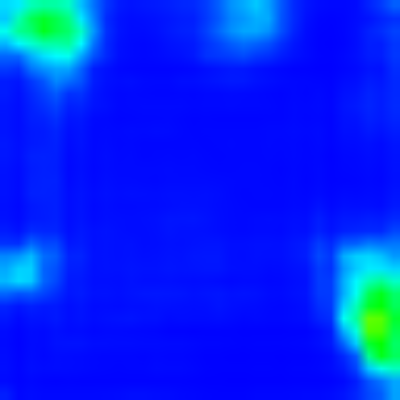}
}
\subfloat[][$\hat{\Omegavect}$ - \emph{Linear Reg.}]{\label{fig:omega_lin_synt}
    \includegraphics[height=0.30\linewidth]{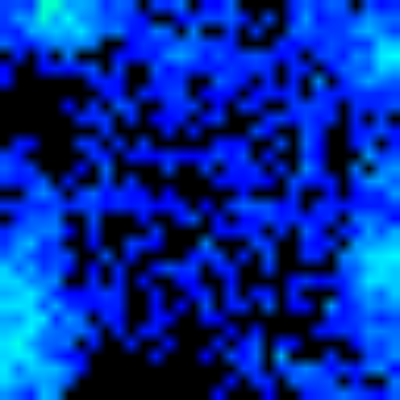}
}\\
\subfloat[][Obj - \emph{Mean-2D-Enc}]{\label{fig:obj_2D_synt}
    \includegraphics[height=0.30\linewidth]{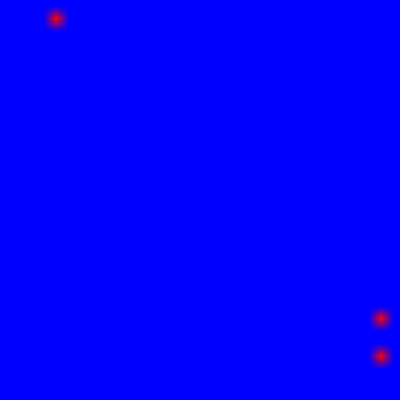}
}
\subfloat[][Obj - \emph{3D/2D U-Net}]{\label{fig:obj_3D_synt}
    \includegraphics[height=0.30\linewidth]{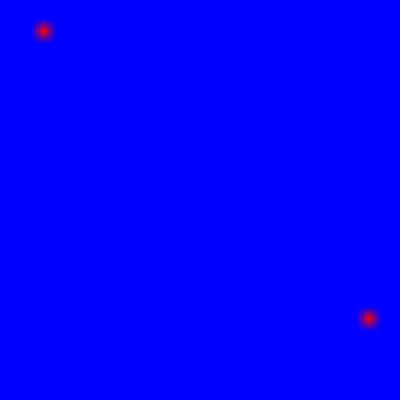}
}
\subfloat[][Obj - \emph{Linear Reg.}]{\label{fig:obj_lin_synt}
    \includegraphics[height=0.30\linewidth]{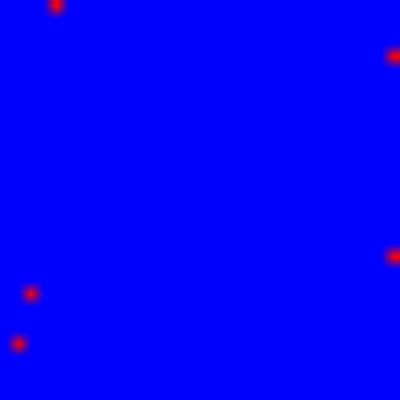}
}

\caption[Hello]{Results of three methods on the \synt sequence from Fig.~\ref{fig:illustration_synt_1}
  \subref{fig:omega_2D_synt}, \subref{fig:omega_3D_synt}, \subref{fig:omega_lin_synt}: Estimates $\hat{\Omegavect}$ of the \synt \emph{object heat-map} $\Omegavect$ from Fig.~\ref{fig:o_synt_1} using three different architectures.
  \subref{fig:obj_2D_synt}, \subref{fig:obj_3D_synt}, \subref{fig:obj_lin_synt}
  : Corresponding objects positions, obtained as the highest local maxima from $\hat{\Omegavect}$.
  Black pixels in
  ~\subref{fig:omega_lin_synt}
  indicate negative values.
}
\label{fig:results}
\end{figure}

\newpage

In Fig.~\ref{fig:results}, the predicted \emph{gaze heat-maps} $\hat{\Omegavect}$ for several learning-based approaches applied on the \synt scenario from Fig.~\ref{fig:illustration_synt_1} are displayed. The architectures \emph{Mean-2D-Enc} and \emph{Linear Reg.} use the average \emph{gaze heat-map}  $\frac{1}{T} \sum_{t=1}^T \Gammavect_t$ as input, whereas \emph{3D/2D U-Net} takes the whole concatenated sequence $\Gammavect_{1:T}$. Contrary to the experiments on the \vern dataset, We observe that the \emph{3D/2D U-Net} yields an \emph{object heat-map} $\hat{\Omegavect}$ closer to the expected one $\Omegavect$ than the other models, and lead to a higher precision. This is consistent with the quantitative results reported in Table~I in the main paper.

\newpage
\subsection{Architectures}

\begin{figure}[!h]
\centering
\subfloat[\emph{Mean-2D-Enc}]{\label{fig:gf-network-mean2D}\includegraphics[width=0.95\linewidth]{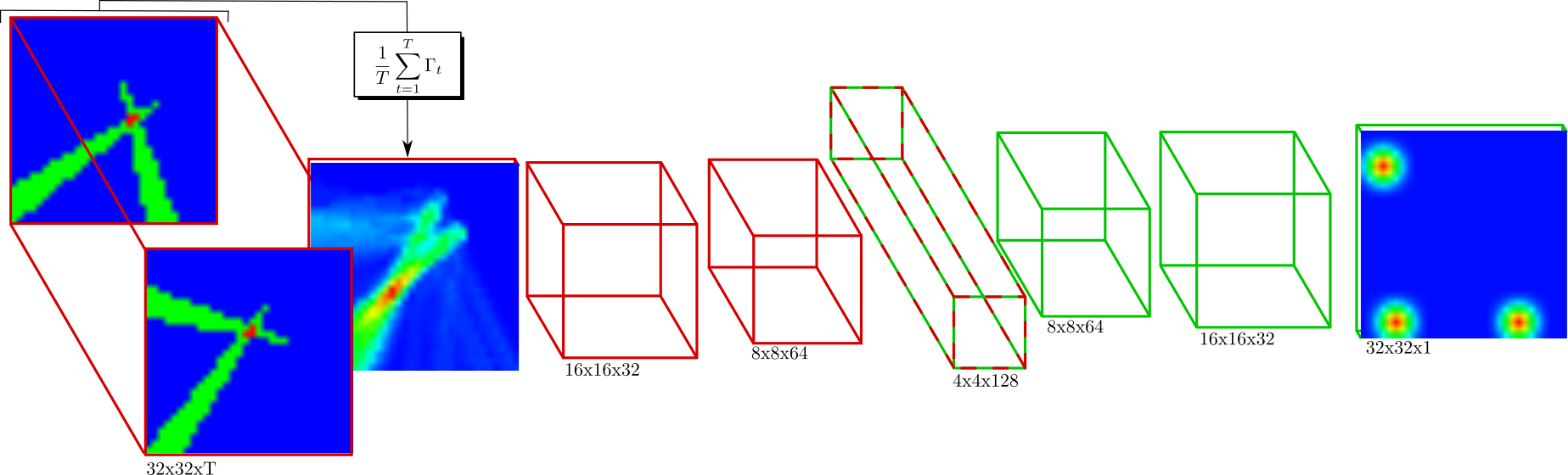}} \\
\subfloat[\emph{2D-Enc}]{\label{fig:gf-network-2D}\includegraphics[width=0.95\linewidth]{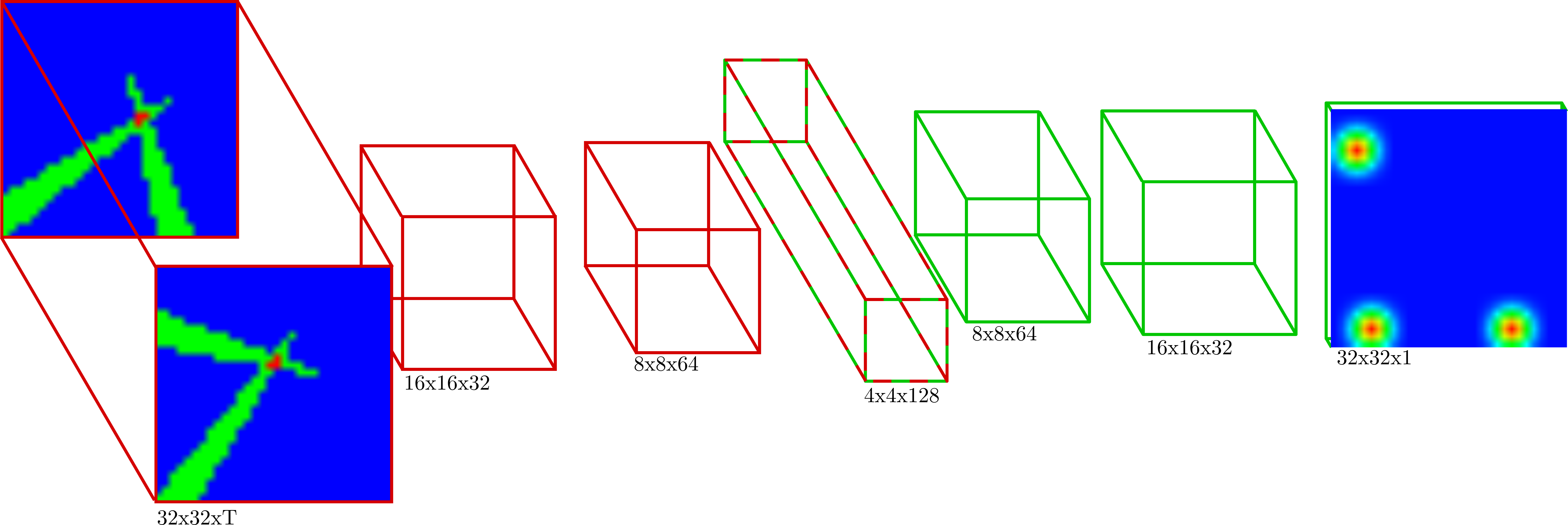}}\\
\subfloat[\emph{3D-Enc}]{\label{fig:gf-network-3D}\includegraphics[width=0.95\linewidth]{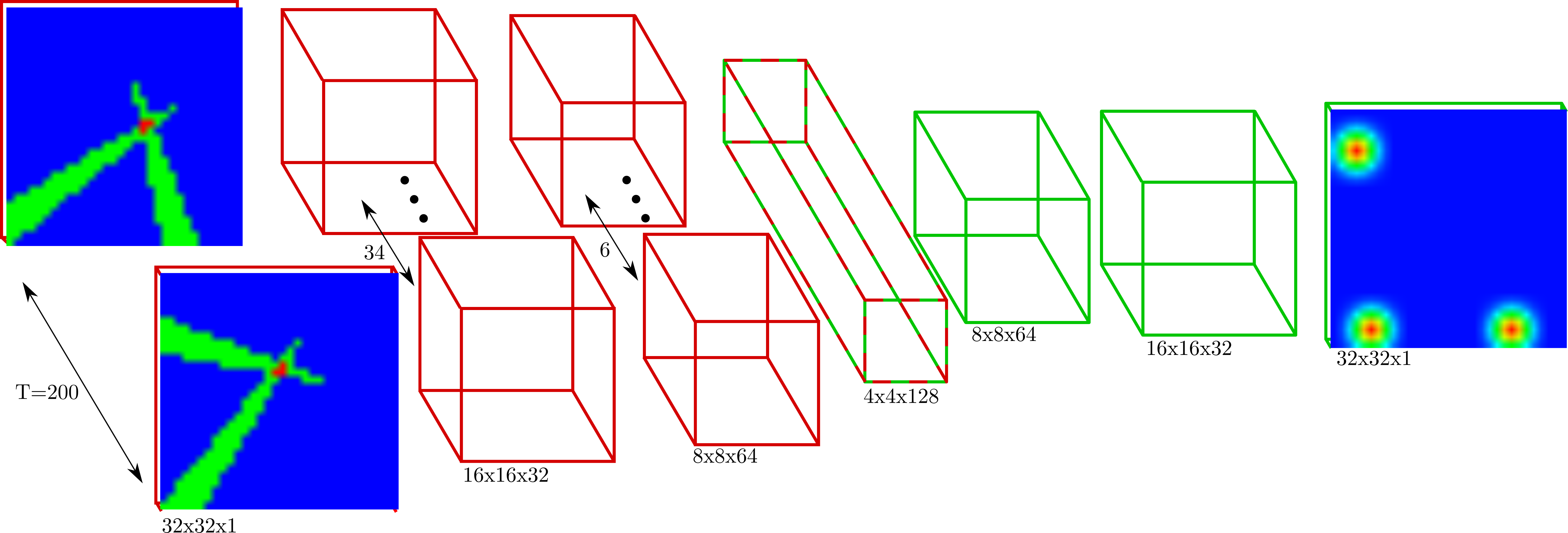}}\\
\subfloat[\emph{3D/2D U-Net}]{\label{fig:gf-network-3DUnet}\includegraphics[width=0.95\linewidth]{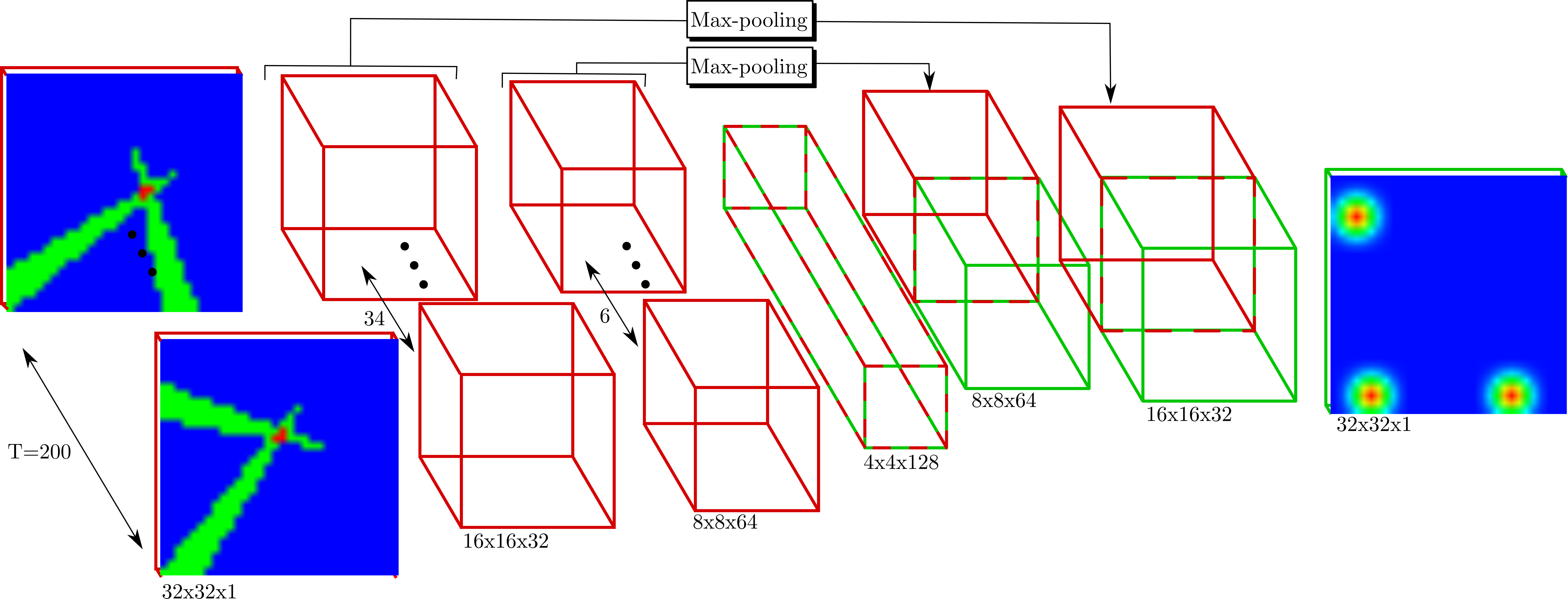}}
\caption[Proposed architectures]{Proposed convolutional encoder/decoder architectures\label{fig:gf-archis}}
\end{figure}

Fig.~\ref{fig:gf-archis} is an illustration of the convolutional encoder/decoder architectures proposed in section III-B of the main paper.

\end{document}

%% file: sec_introduction.tex
\section{Introduction}
\label{sec:gf-intro}

\emph{Gaze following} is the ability to intuit the region of space that an observer is looking at. Humans learn this skill during infancy~
\cite{baldwin1995understanding}, and use it very frequently in many social activities~\cite{land2009looking}. An accurate estimation of where one or several persons look has an enormous potential in order to determine which are the objects of interest in a scene, predict the actions and movements of the participants and, in general terms, advance towards a better visual scene understanding. It has applications in various fields such as human-robot interaction~\cite{schauerte2014look, domhof2015multimodal, lathuiliere2019prl}, or action recognition~\cite{Wei2018Where}. However, automatically estimating the visual region of attention remains an open challenge, particularly when the gaze target is not visible within the field of view.

\begin{figure*}
\centering
\subfloat[Gaze following~\cite{Recasens2015}]{\label{fig:gf-gazeF}\includegraphics[height=0.215\linewidth]{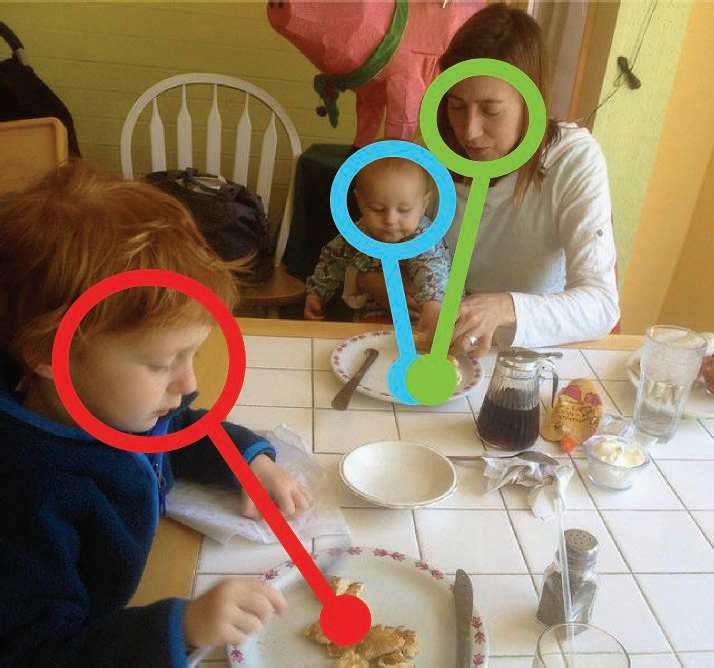}}
\subfloat[Visual Focus of Attention~\cite{Masse2017}]{\label{fig:gf-VFOA}\includegraphics[height=0.175\linewidth]{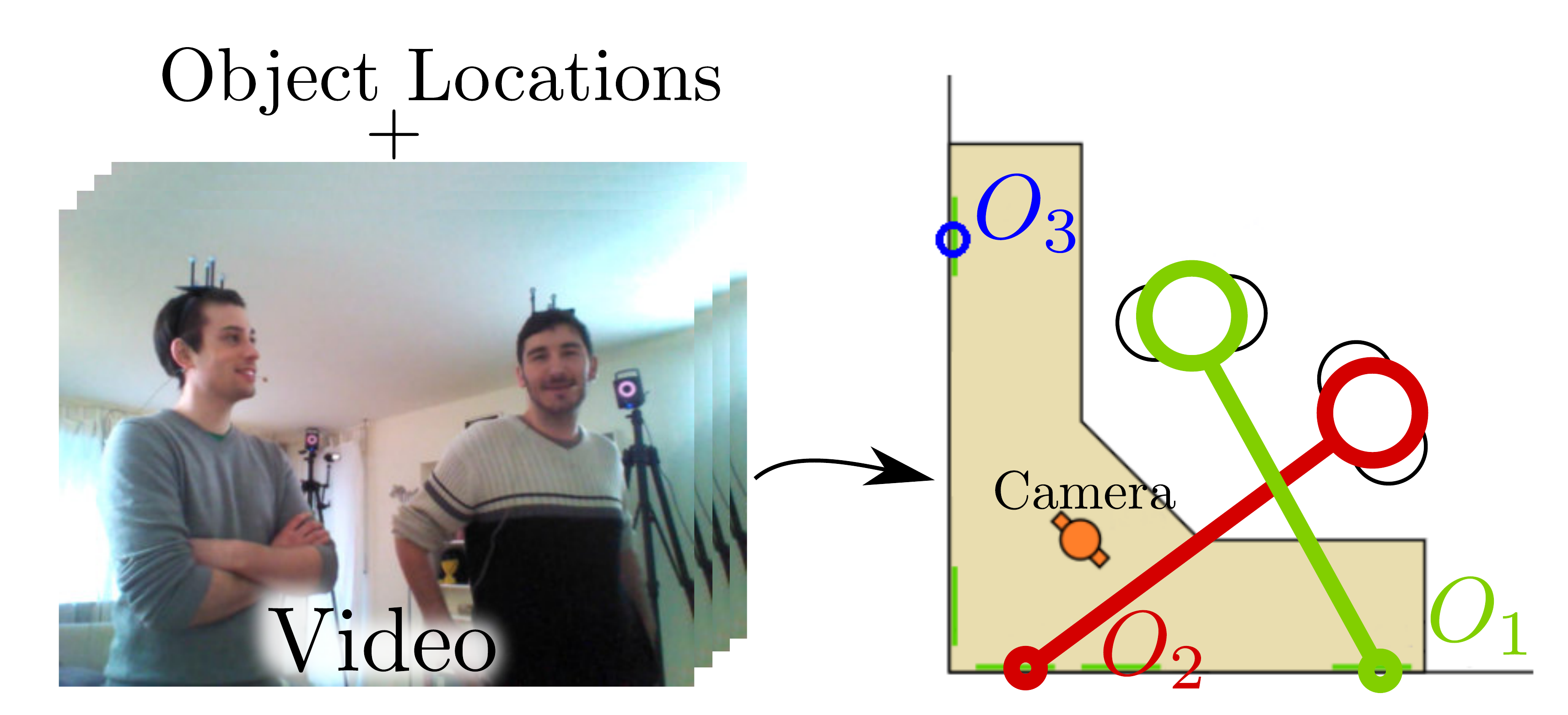}}
\subfloat[Extended Gaze following]{\label{fig:gf-OFVOD}\includegraphics[height=0.175\linewidth]{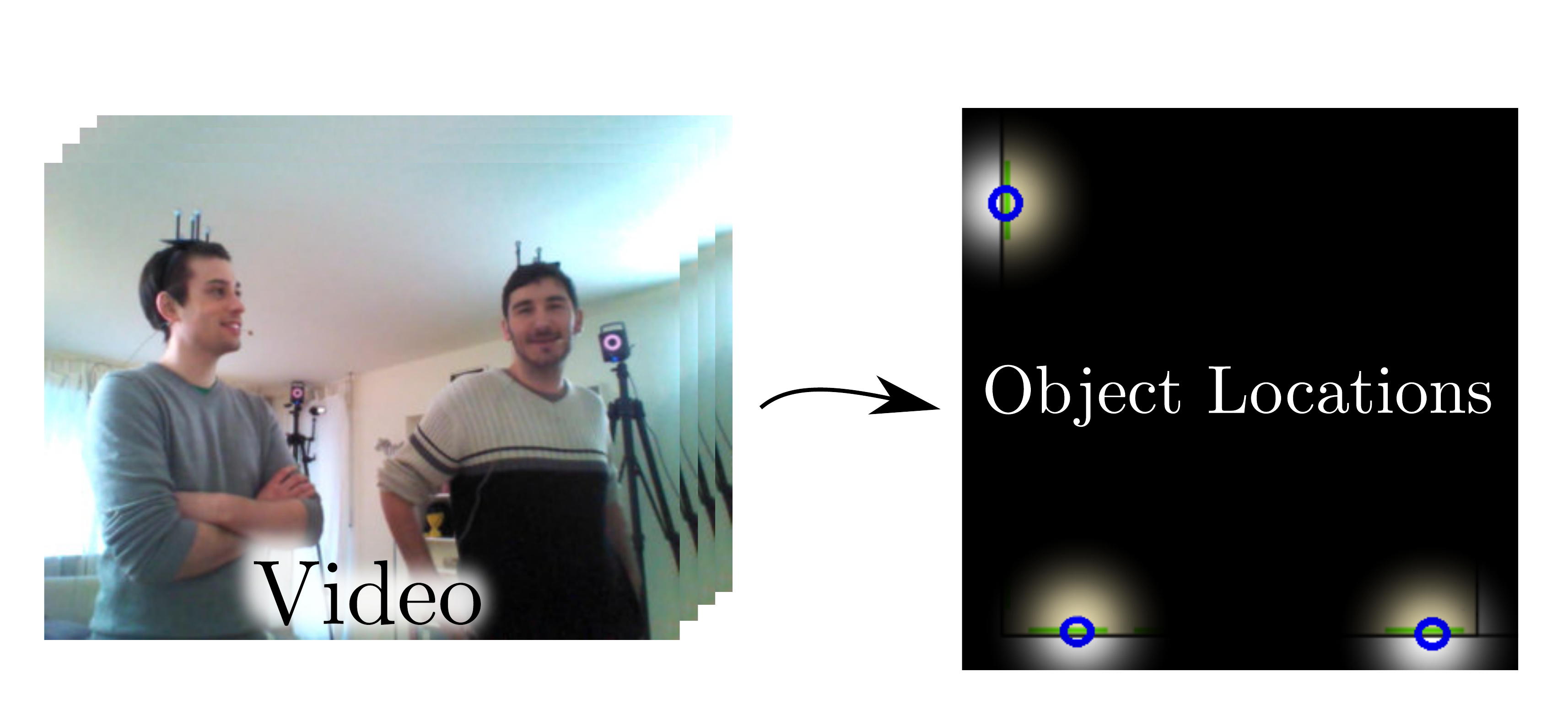}}

\caption[A comparison of gaze-related computer vision problems]{
  A comparison of gaze-related computer vision problems. In the standard formulation of gaze following (a), the problem consists in localizing the objects that people are likely to be looking at (and both observer and objects are visible in the input image). Visual focus of attention estimation (b) consists in associating which person is looking at what object at a certain moment (considering that the objects locations are known). In extended gaze following (or visual regions of attention detection) (c), we aim at localizing objects of interest even if they are not visible in the video image. }
\label{fig:gf-gaze-cv-problems}
\end{figure*}

This paper addresses the detection of visual regions of attention, which are expected to contain objects of interest. People in a video generally either look at other people or at an object of interest. Such an object can be indistinctly located inside or outside the current image. In the standard \emph{gaze-following} problem, addressed \eg in \cite{Recasens2015}, both the observer and the targeted object are within the same image. An example is provided on Fig.~\subref*{fig:gf-gazeF}.
 This is related to -- but significantly different from -- estimating the \emph{visual focus of attention} \ie whom or what a person is looking at~\cite{Masse2017}. In this case, object locations are known, but potentially non-visible (occluded or outside the field of view, see Fig.~\subref*{fig:gf-VFOA}). However, in a general setting, an object may not be visible within the image, and its location is most probably unknown.
All the more in a social interaction, an object is not ``of interest'' until people actually start paying attention to it. In this paper, we deal with \emph{extended gaze following} in videos, see Fig.~\subref*{fig:gf-OFVOD}, meaning that we tackle the more general problem of predicting the location of objects of interest whose number and locations are not known a priori, and that are not necessarily visible.

Our method takes as input a video sequence containing a group of people, and outputs a set of estimated locations for the objects of interest. This work makes the assumption that objects do not move across the video sequence. As in~\cite{Masse2017, mukherjee2015deep}, we propose to use the head orientation as a strong cue for the gaze direction. 
The pipeline is illustrated in Fig.~\ref{fig:gf-outline}.

The contribution of this paper is threefold. First, we propose a novel formalism for embedding the spatial representation of directions of interest and regions of attention. They are modeled as a top-view heat-map, \ie a discrete grid of spatial regions from a top-view perspective. Contrary to the majority of previous work, this formalism is not limited to representing locations within the field of view. Second, we propose several convolutional encoder/decoder neural architectures that learn to predict object locations from head poses in our proposed embedding, and we compare them with several baselines inspired from earlier work. Third, since a large amount of data are required to train a deep neural network, we propose an algorithm based on a generative probabilistic framework that can sample an unlimited number of synthetic conversational scenarios, involving people and objects of interest. The method has been tested both on synthetic data and on a publicly available dataset.

The remainder of this paper is organized as follows. The state of the art is presented in Section~\ref{sec:gf-related}. Then, the details of the proposed heat-map representations and neural network architectures are respectively given in Sections~\ref{sec:HM} and~\ref{sec:gf-inference}. The synthetic data generation process is described in Section~\ref{sec:synthetic}. The Section~\ref{sec:gf-expe} is dedicated to experimental results, both on synthetic and real data. To conclude, Section~\ref{sec:gf-concl} discuss the perspectives and limitations of this work.

%% file: sec_related.tex
\section{Related work}
\label{sec:gf-related}


\emph{Gaze following}, or more generally any problem based on the visual attention of a person within an image, is intrinsically based on estimating the gaze direction. In practice, estimating the gaze direction is a complicated problem that still requires to compromise between being precise and non-invasive. When precision is crucial, a head-mounted system, \eg~\cite{Hong2012}, can provide very accurate gaze direction. However, it cannot be used in a natural scenario since it requires a specific setup. Since the head-mounted system is visible to all participants, it may bias what would be the nature of social behaviour. Another issue is that the head-mounted system can hardly be used to annotate training data since the system would appear in the images recorded by external cameras and, therefore, real environment images that are recorded without head-mounted system would differ from the training set images~\cite{Fischer_2018_ECCV}.
On the other hand, estimating gaze direction from remote camera images is a difficult task, with non-frontal faces, or eyelid occlusions~\cite{Hansen2010}. Moreover, since it is difficult to obtain  gaze annotations in scenarios where people can move freely, most learning-based methods are trained on extremely simplified setups. For instance, in~\cite{Krafka2016} and~\cite{zhang2015appearance}, subjects were asked to fixate a region on the screen of a camera-equipped device. Alternatively, in unconstrained settings, the head pose is highly correlated with the gaze direction, and the former can be used as an approximation for the latter~\cite{mukherjee2015deep, Masse2017}.



Finding objects of interest generally requires to analyze the visual field of view and look for highly contrasting regions. Indeed, an object or a person is likely to look different from the background, thus highly contrasting regions have higher chance of containing something interesting. This approach, similar to the human brain pipeline~\cite{treue2003visual}, is known in computer vision as \emph{saliency}~\cite{itti2001computational,rudoy2013learning,parks2015augmented,wang2018deep}, where a salient region is one that attracts the visual attention of an observer. 
In the context of gaze following, the goal is to find regions that are salient, \ie that attracts gaze, from another point of view. However, a salient region is most likely salient from most points of view. Based on this remark,~\cite{Recasens2015} combines a saliency model with a gaze direction model to find salient objects at the intersection of the image and the person's field of view. The attention predictor in~\cite{Wei2018Where} also uses both saliency and gaze. By combining multiple gaze directions, \cite{fan2018inferring} estimates shared attention of multiple people, but still within the image. In~\cite{recasens2017following}, the authors further investigate this problem based on the idea that the gaze target of a person inside a video may be visible in another video frame. Their method still relies on a saliency model. Recently, \cite{Chong_2018_ECCV} uses a similar combination of gaze and saliency but is also able to predict whether the object of attention lies within the image or not.
Finally, \cite{schauerte2014look, domhof2015multimodal} merge the problems of saliency and gaze following in the context of human-robot interaction. Indeed, the robot is both an active member of the scenario, and an observer behind the camera. Both papers are based on saliency and gaze direction, as well as additional data such as pointing gesture and speech.
However, all works based on saliency require that the object of interest lies within the field of view. By contrast, we wish to be able to locate out-of-view objects; therefore, we cannot rely on this category of methods.

\begin{figure*}[t!]
 \centering
 \includegraphics[width=\linewidth]{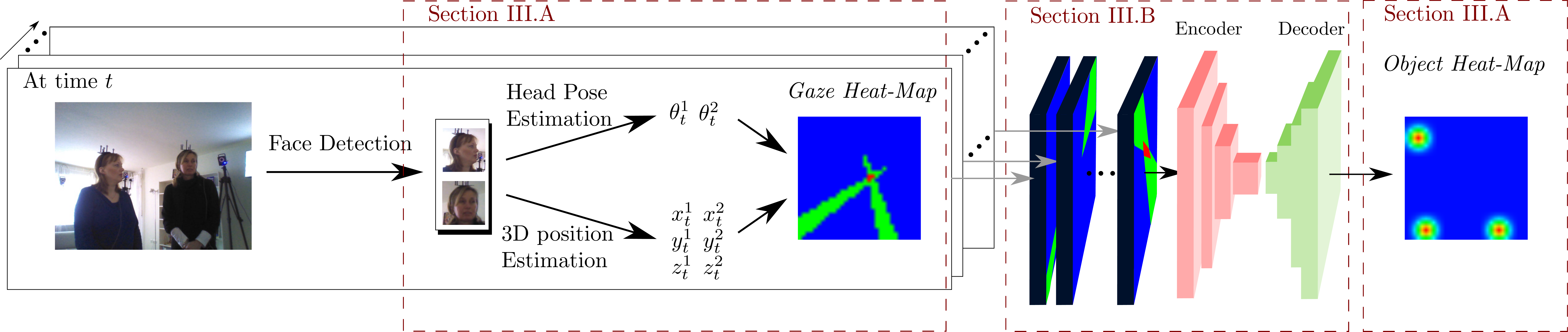} 
 \caption[Outline of the proposed model]{
   Outline of the proposed model. For every frame and detected face, orientation and 3D location are estimated, and both sources of information are combined to obtain a top-view representation of the scene encoded in a heat-map. The sequence of heat-maps is then given to a neural network with an encoder/decoder architecture. The network outputs a heat-map that predicts the position of the objects of interest in the top-view domain.
  }
 \label{fig:gf-outline}
\end{figure*}

Apart from saliency-based gaze following, a few other methods have been published, addressing the gaze-following problem in the 3D space instead of the 2D image plane.
\cite{soo2015social} proposes to estimate 3D regions of attention using only the location of people. They model social group structures that constrain the set of candidate locations. In this framework, they learn to locate regions of attention independently of visual saliency. Their method only needs people locations and can work in adversarial scenarios, using only spatial data from first person cameras. However, it fails when some people are undetected and the group structures are wrongly estimated, or when a person is isolated and should not be integrated into a group structure.
By contrast, both~\cite{Cohen2012} and~\cite{Brau_2018_ECCV} independently propose to use 3D intersection of gazes in a probabilistic framework to estimate locations of objects of interest, possibly outside the camera field of view. The methods achieve good levels of performance -- even though~\cite{Cohen2012} lacks quantitative evaluation. 
In both cases, no training data have been used. Each method is designed with strong geometric assumptions so that location inference can be performed without any prior learning phase. At the time this article was written, the data on which the methods have been tested were not released yet for comparison.

In this paper, we combine a learning-based model with a geometric formulation to address the gaze-following problem, without the restriction of being limited to the image plane. Only very few works exist in this direction~\cite{soo2015social, Cohen2012, Brau_2018_ECCV}, and all employ strong social or geometric assumptions.

%% file: sec_model.tex
\section{Deep Learning for Extended Gaze Following}

We note $N_t$ the number of persons at time $t \in \{1\dots T\}$. For each person, we suppose that we can estimate its corresponding 3D head location $[x^{n}_t,y^{n}_t,z^{n}_t]^\top$, and head orientation $[\phi_{t}^n, \theta_{t}^n]^\top$ for person $n \in \{1\dots N_t\}$ in a common scene-centered coordinate frame. However, we additionally choose to drop the z-coordinate (the height) and the head tilt angle as in \cite{Cohen2012}, projecting every object and every person in the same horizontal plane. As we will see later, this simplification drastically reduces the complexity of the model while still representing plausible scenarios. In addition, the tilt angle is commonly the one estimated with the largest mean absolute error \cite{lathuiliere17}.
In the remaining of the paper, the term \emph{position} refers to 2D coordinates $\xmat_t^n = [x_t^n, y_t^n]^\top$ in the horizontal plane (top-view perspective), and \emph{head orientation} refers to the head pan angle $\phi_t^n$.

As mentioned before, we decided to use heat-map embeddings. The reasons for this are multiple.
First, the exact number of people and objects is not known a priori and may vary within and between video sequences. Heat-map structures are independent of the number of participants (people and objects of interest). Additionally, the problem addressed is fundamentally geometric, and heat-maps intrinsically encode the geometry of the scene. Moreover, convolutional neural networks are able to efficiently extract this structured information in order to obtain a descriptive input representation. A drawback of the heat-map representation is the difficulty to predict an object outside the modeled area. Nevertheless, for indoor scenarios, the area containing the objects is bounded. It is then possible to adapt the heat-map size for the current setup and train the model using scaled simulated scenarios (see Section \ref{sec:synthetic}). For all these reasons, we employ heat-map embeddings to model the geometry of the scene.

\subsection{Heat-Map Representation}
\label{sec:HM}

We propose several heat-map representations of the scene from a top-view perspective. The scene is discretized into a 2D grid of dimension $S_U\times S_V$. 
Each position in the scene $\xmat=(x,y)$ is associated to a grid cell $\pmat=(u,v) \in \{1\dots S_U\}\times \{1\dots S_V\}$. As stated previously, $\xmat$ is bounded in both dimensions: $x \in [x_{min}, x_{max}]$ and $y \in [y_{min}, y_{max}]$. With these notations, $\pmat=(u,v)$ is obtained from $\xmat$ as
\begin{equation}
  \begin{cases}
    u &= \lceil S_U \times \frac{x-x_{min}}{x_{max}-x_{min}} \rceil\\
    v &= \lceil S_V \times \frac{y-y_{min}}{y_{max}-y_{min}} \rceil\\
  \end{cases}
\end{equation}
where $\lceil \cdot \rceil$ is the \emph{ceiling} function.
The grid cell associated to  $\xmat^n_t=(x^{n}_t,y^{n}_t)$, the position of a person $n$ at time $t$, is  $\pmat^n_t$.

In this formalism, a heat-map $\Lambdavect$ is a 2D map of $S_U \times S_V$ elements that attaches to each cell $\pmat$ of the grid a value $\Lambdavect(\pmat)$ between $0$ and $1$. The meaning of this value depends on what the heat-map represents. In this paper, there are two different categories of heat-map. First, a \emph{gaze heat-map} $\Gammavect$ is an embedding for head pose information. A value close to one indicates a region of space consistently situated in front of someone's head. Second, an \emph{object heat-map} $\Omegavect$ embeds the likelihood for each region to contain an object of interest.

\begin{figure}

\centering     
\subfloat[][Camera Image at $t=10$]{\label{fig:camera10}
    \includegraphics[height=0.23\linewidth]{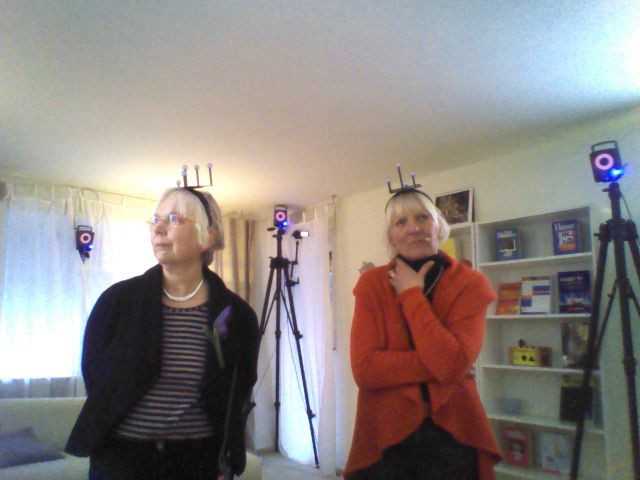}
} \hspace{0.2cm}
\subfloat[][\emph{Gaze heat-map} $\Gammavect_{10}$]{\label{fig:hm10}
    \includegraphics[height=0.23\linewidth]{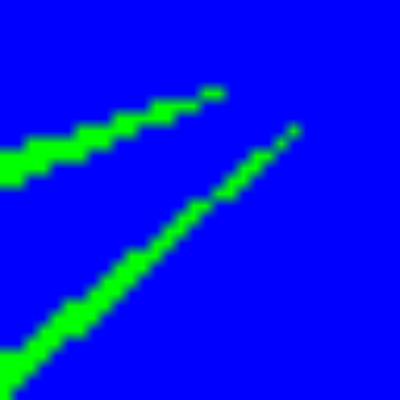}
} \hspace{0.4cm}
\subfloat[][Mean \emph{gaze heat-map} $\Gammavect$]{\label{fig:mean_hm} 
    \includegraphics[height=0.23\linewidth]{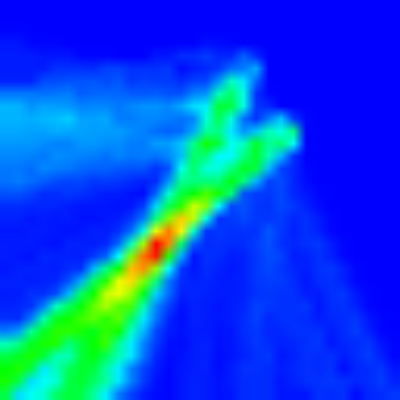}
} \\
\subfloat[][Camera Image at $t=30$]{\label{fig:camera30}
  \includegraphics[height=0.23\linewidth]{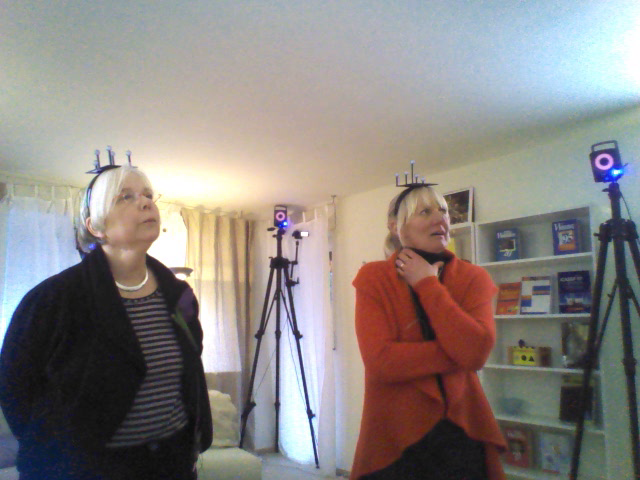}
} \hspace{0.2cm}
\subfloat[][\emph{Gaze heat-map} $\Gammavect_{30}$]{\label{fig:hm30}
  \includegraphics[height=0.23\linewidth]{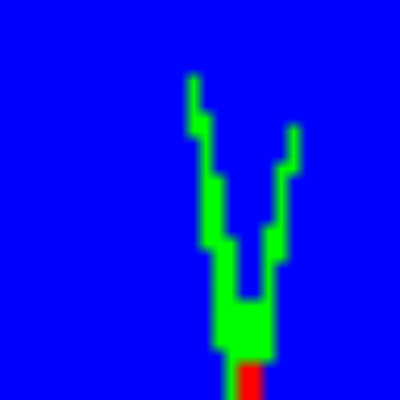}
} \hspace{0.4cm}
\subfloat[][Object positions]{\label{fig:objects}
    \includegraphics[height=0.23\linewidth]{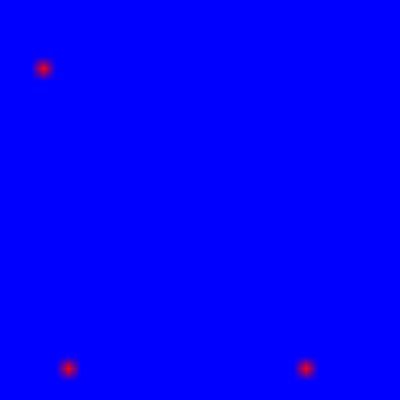}
} \\
\subfloat[][Camera Image at $t=80$]{\label{fig:camera80}
  \includegraphics[height=0.23\linewidth]{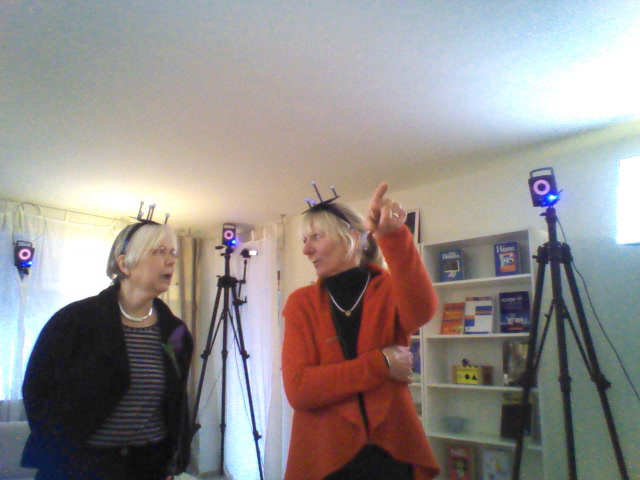}
} \hspace{0.2cm}
\subfloat[][\emph{Gaze heat-map} $\Gammavect_{80}$]{\label{fig:hm80}
  \includegraphics[height=0.23\linewidth]{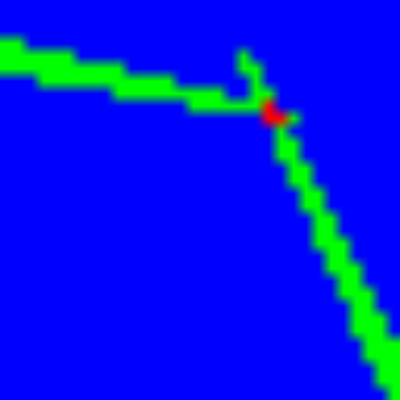}
} \hspace{0.4cm}
\subfloat[][\emph{Object heat-map} $\Omega$]{\label{fig:o_hm}
    \includegraphics[height=0.23\linewidth]{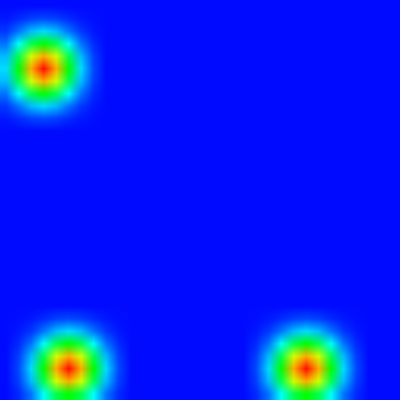}
}

\caption[Illustration of the heat-map representations on a \vern sequence]{
  Illustration of the heat-map representations using a sequence extracted from the \vern dataset~\cite{jayagopi2013vernissage}. The camera is located close to the bottom left corner of the \emph{gaze heat-maps}. Heat-map colors range from blue to red to indicate number from $0$ to $1$.
  \subref{fig:camera10}, \subref{fig:camera30}, \subref{fig:camera80}: camera images. 
  \subref{fig:hm10}, \subref{fig:hm30}, \subref{fig:hm80}: corresponding \emph{gaze heat-maps}. Cone origins in the \emph{gaze heat-maps} indicate people positions; cone axes represent head orientations.
  \subref{fig:mean_hm}: mean \emph{Gaze heat-maps} over the sequence. 
  The object ground truth is represented in the heat-map coordinate frame \subref{fig:objects}. This provides the ground truth \emph{Object heat-map} \subref{fig:o_hm} used for training and \emph{MSE} evaluation.
}
\label{fig:illustration}
\end{figure}

\paragraph{\emph{Gaze heat-map} representation $\Gammavect$}
Motivated by the use of cones for modeling the dependency between head pose and gaze \cite{Marin-Jimenez2014}, we compute a heat-map  $\Gammavect^{n}_t \in [0,1]^{S_U\times S_V}$ for each person $n \in \{1\dots N_t\}$ by considering a cone whose axis is the direction spanned by the head pan angle $\phi_t^n$. Formally, the value of $\Gammavect^{n}_t$ at any grid cell $\pmat$ is given by:
\begin{equation}
  \label{eq:gazemap}
  \Gammavect^{n}_t(\pmat) =
  \begin{cases}
    1 &\quad\text{if }|\phi(\pmat)-\phi^{n}_t|<\epsilon\\
    0 &\quad\text{otherwise} \\
  \end{cases}
\end{equation}
where $\phi(\pmat)$ is the angle corresponding to the direction of vector $\overrightarrow{\pmat^n_t\pmat}$. The parameter $\epsilon$ controls the aperture of the cone. 
We obtain the \emph{gaze heat-map} illustrated in Fig.~\subref*{fig:hm10}, \subref*{fig:hm30} and \subref*{fig:hm80}:
\begin{equation}
  \Gammavect_t= \frac{1}{N_t} \sum_{n=1}^{N_t} \Gammavect^{n}_t.
\end{equation}
It is sometimes useful to aggregate the \emph{gaze heat-maps} through time into a mean \emph{gaze heat-map} (see Fig.~\subref*{fig:mean_hm}) to have an compact representation of the scenario:
\begin{equation}
  \Gammavect=  \frac{1}{T} \sum_{t=1}^T \Gammavect_t.
\end{equation}

\paragraph{\emph{Object heat-map} $\Omegavect$}
Considering a scenario with $M$ objects (\eg Fig.~\subref*{fig:objects}), we compute a heat-map $\Omegavect \in [0,1]^{S_U\times S_V}$ (Fig.~\subref*{fig:o_hm}) whose value at grid cell $\pmat$ is given by:
\begin{equation}
  \Omegavect(\pmat) =\max_{1\leq m\leq M}\exp\left(-\frac{||\pmat-\pmat^m_{obj}||_2^2}{2\sigma_\Omega^2}\right)
\end{equation}
where $\pmat^m_{obj}$ is the grid cell corresponding to the scene position of the $m^{th}$ object. The variance $\sigma_\Omega$ controls the spread of the peaks. As objects do not move, $\Omegavect$ remains constant during a scenario.

Now, let us suppose we have been able to obtain an estimate $\hat{\Omegavect}$ of $\Omegavect$ from $\Gammavect_1 \dots \Gammavect_T$.
Finally, to obtain an actual list of object positions, we extract the local maxima from $\hat{\Omegavect}$ and discard local maxima that are too low compared to the global maximum. More precisely, given a candidate position $\pmat_C$, a neighborhood of this position $\mathpzc{N}(\pmat_C)$ and a shrinking function $\alpha(\cdot)$ such that $\alpha(x)\leq x$, we consider that $\pmat_C$ contains an object if
\begin{equation}
  \label{eq:NMS}
  \pmat_C = \argmax_{\pmat \in \mathpzc{N}(\pmat_C)} \hat{\Omegavect}(\pmat) \qquad \text{and} \qquad  \hat{\Omegavect}(\pmat_C) \ge \alpha \left( \max_{\pmat}  \hat{\Omegavect}(\pmat) \right).\\
\end{equation}

The section~\ref{sec:gf-inference} below is dedicated to propose a neural network that learns to predict an estimate $\hat{\Omegavect}$ of the \emph{object heat-map} from the set of \emph{gaze heat-maps} $\Gammavect_1\dots\Gammavect_T$.

\subsection{Object heat-map inference}
\label{sec:gf-inference}


Now, we address the problem of estimating $\hat{\Omegavect}$, on which the local maxima detection algorithm can be run. We propose several baselines with justification for their relevance. Then, we present our architectures based on convolutional encoder/decoder.

\paragraph{Heuristics without learning}
First, we propose two heuristics with no training. The local maxima detection is performed directly on a combination of gaze heat-maps. Indeed, the regions that are activated (close to one) in multiple gaze heat-maps are consistently in front of someone's head and have a high chance of containing an object. Previous works~\cite{Cohen2012, Marin-Jimenez2014} already used geometric features based on cone intersections. The heuristics are as follow.
\begin{itemize}
\item \emph{Cone}: The local maxima extraction is performed directly on the \emph{mean gaze heat-map} $\Gammavect = \frac{1}{T}\sum_{t=1}^{T}\Gammavect_t$.
\item \emph{Intersect}: We define a \emph{gaze intersection heat-map} $\Gammavect_t^{inter}$ per time frame, by setting regions to one only if they are at the intersection of multiple cones. More formally,
\begin{equation}
  \Gammavect_t^{inter} (\pmat) =
  \begin{cases}
    1 & \text{if} \sum_{n=1}^{N_t} \Gammavect_t^n (\pmat) \geq 2 \\
    0 & \text{otherwise}
\end{cases}
\end{equation}
The local maxima extraction is performed on $\Gammavect^{inter} = \frac{1}{T} \sum_{t=1}^T \Gammavect_t^{inter}$.
\end{itemize}

\paragraph{Learning-based Baselines}
We define some simple regression models. They learn a regression from the \emph{mean gaze heat-map} $\Gammavect = \frac{1}{T}\sum_{t=1}^{T}\Gammavect_t$ to the \emph{Object heat-map} $\hat{\Omegavect}$. These models consider the input and output as flattened vectors of $S_U\times S_V$ components.
\begin{itemize}
\item  \emph{Linear Reg.}: We learn a linear regression model from $\Gammavect$ to $\hat{\Omegavect}$. Interestingly, the output of a linear regression is not constrained to lie between $0$ and $1$, contrary to the definition of $\Omegavect$. The local maxima extraction is performed after $\hat{\Omegavect}$ has been rescaled in $[0,1]$.
\item \emph{d-FC}: The regression is performed on $\Gammavect$ by a network composed of $d\in\{1,3\}$  fully connected hidden layers of $S_U\times S_V$ units, with ReLU activations. The last hidden layer is fully connected to the output \emph{object heat-map} with sigmoid activations.
\end{itemize}

\paragraph{Encoder/Decoder Architectures} They have been used for many computer vision tasks where the goal is to perform a regression between high dimensional spaces \cite{isola2017image,badrinarayanan2017segnet}. Such architectures are composed of two sub-networks, where the first reduces the spatial resolution of the input to obtain a compact description of it, and the second alternates between up-sampling and fully-connected layers until recovering a high dimensional output. In our particular problem, we use convolutional layers instead of fully-connected layers to model the spatial connections. Moreover, as the input is a sequence, several encoder architectures can be employed. We propose to use a decoder composed of three successive up-sampling and convolutional layers with $3 \times 3$ kernels. The last convolution layer of the decoder employs sigmoid activations. The whole network is trained employing the Mean Squared Error (MSE) loss. We propose the four following architectures that represent a progressively increasing complexity. Graphical representations of the proposed networks are given in the supplementary material\footnote{see \textit{https://team.inria.fr/perception/research/extended-gaze-following}}.
\begin{itemize}
\item\emph{Mean-2D-Enc}: 
This is the simplest model. We use the \emph{mean gaze heat-maps} $\Gammavect$ as in the baselines. It is fed to a standard 2D convolutional encoder composed of three successive convolutional and down-sampling layers.
\item\emph{2D-Enc}: 
In this model, we consider that time plays the role of the color-axis in standard 2D convolutions. $\Gammavect_1\dots\Gammavect_T$ are concatenated along the third dimension to obtain the \emph{sequence gaze heat-map} $\Gammavect_{1:T}$. Therefore, the first layer kernels have as dimension $3\times3\times T$ instead of $3\times3\times1$ like in \emph{Mean-2D-Enc}.
\item\emph{3D-Enc}: 
 Inspired by \cite{ji20133d}, that shows that 3D convolutions are able to extract reliable features from both the spatial and the temporal dimensions, we propose a 3D-Encoder network on $\Gammavect_{1:T}$. By performing 3D convolutions, the model can capture orientation changes and people motion in successive frames. The time dimension is reduced, from $T$ to $1$ after three convolutional and max-pooling layers, before feeding it to the 2D-Decoder.
\item\emph{3D/2D U-Net}: 
  This variant of the \emph{3D-Enc} architecture is inspired from the U-Net architecture \cite{ronneberger2015u}. In our specific case, since we have a 3D encoder, we need to squeeze the time dimension. To do so, we combine over time the feature maps of the encoder with max-pooling, before concatenation to the decoder.
\end{itemize}

%% file: sec_data.tex
\section{Synthetic Scenario Generation for Network Training}
\label{sec:synthetic}

A large amount of data is required to train deep networks. Unfortunately, obtaining such a dataset is difficult, since, in practice, we would need to know the true object locations for every sequence. For instance, in the \vern dataset~\cite{jayagopi2013vernissage}, objects outside the field of view have been annotated employing infrared cameras. This setting is well-suited for our problem but it would be difficult to obtain a sufficiently large and diverse dataset of object locations to train deep networks. Consequently, Vernissage is used only to test our model and not to train it.

To face this issue, we propose to use synthetically generated data. More precisely, we simulate scenarios involving people and objects, and generate their corresponding input sequences and associated true object locations. We define a probabilistic model that relates the object 2D positions and the head poses, and generate samples according to the underlying distribution. We now aim at generating a scenario of length $T$ involving a constant number $N$ of people with respective positions $\xmat_{1:T}^n$ and orientations $\phimat_{1:T}^n$, given $1<n<N$; and $M$ objects located at the positions $\xmat^m_{obj}, 1<m<M$. To this aim, we define the joint distribution $P(\phimat_{1:T}^{1:N}, \xmat_{1:T}^{1:N},\xmat^{1:M}_{obj})$ considering the following factorization:
\begin{multline}
  \underbrace{P(\phimat_{1:T}^{1:N}| \xmat_{1:T}^{1:N},\xmat^{1:M}_{obj})}_{\substack{\text{Head orientation} \\\text{ distribution}}}\times\underbrace{P(\xmat_{1:T}^{1:N}|\xmat^{1:M}_{obj})}_{\substack{\text{People motion } \\\text{distribution}}}\times\underbrace{P(\xmat^{1:M}_{obj})}_{\substack{\text{Object position} \\\text{distribution}}}
  \label{eq:sample}
\end{multline}

The \emph{object position distribution} $P(\xmat^{1:M}_{obj})$ is based on a uniform distribution within the top-view grid, since we want to have a high variety of settings. However, some settings are even too difficult  for a human to distinguish between objects. For this reason, the generator can choose to resample an object under two criteria. First, the closest two objects are from each other, the highest the chance one of them is resampled. Therefore, we impose that objects have a minimal physical size and that two objects cannot be one above the other. 
Second, objects too far from the heat-map edges also have a high chance of being resampled. In many scenarios, objects of interest tend to be close to the walls, \eg posters, computer screens, paintings in a museum. This tends to reduce the number of ambiguous cases in which several objects are aligned from the point of view of someone.

Importantly, in a human-robot interaction scenario, people may look at the robot, but we want to avoid our model to predict the presence of an object at the robot camera position.
Therefore, as the camera position $\xmat_{camera}$ is known, we propose to add a blank object at the corresponding grid cell $\pmat_{camera}$ in all sequences. The blank object behave like normal objects -- constant position, can be gazed at -- but does not appear in the \emph{object heat-map} at training time and thus should be ignored at prediction time. Also, it cannot be resampled while generating the objects.

Concerning the \emph{people motion distribution}, $P(\xmat_{1:T}^{1:N}|\xmat^{1:M}_{obj})$, we describe first how the initial positions $\xmat_1^{1:N}$ are sampled, and then how each $\xmat_{t+1}^n$ is sampled iteratively from $\xmat_{t}^n$. 
First, the initial positions of people are obtained similarly to object positions. Namely, they are sampled uniformly within the boundaries, and can be resampled when too close to an object, another person, or (contrary to objects) too close to the edges.
Concerning the motion, we consider that people can either stay still for a random period of time, or move linearly short distances. 
In practice, there is a high probability that the person stay still $\xmat_{t+1}^n = \xmat_t^n$. Otherwise, $\xmat_{t+\tau}^n$ is sampled from a normal distribution centered on $\xmat_t^n$, and possibly resampled as long as $\xmat_{t+\tau}^n$ is outside the boundaries or too close to another target. In the latter case, $\xmat_{t+1}^n\dots \xmat_{t+\tau-1}^n$ are linearly interpolated.

Finally, for the \emph{head orientation distribution}, we define a probabilistic model inspired by~\cite{Masse2017}, where the authors propose a method to estimate the visual focus of attention of multiple people by applying Bayesian inference on a generative model. 
In this probabilistic model, the head orientation dynamics are explained by some latent variables, \eg gaze direction.
For more details, see~\cite{Masse2017}. In our case, we propose to sample the latent variables over time, then sample the head orientation $\phi_t^n$ given the latent variables.
  This model is well-suited for our sampling task since multiple situations may occur, \eg mutual gaze or joint attention, that are treated differently by their temporal formulation. Moreover, it takes into account the discrepancy between head pose and gaze direction, and the network can learn this difference because it is modeled at training time.


\begin{figure}
\centering     
\subfloat[][\emph{Object heat-map}]{\label{fig:o_synt_1}
    \includegraphics[height=0.27\linewidth]{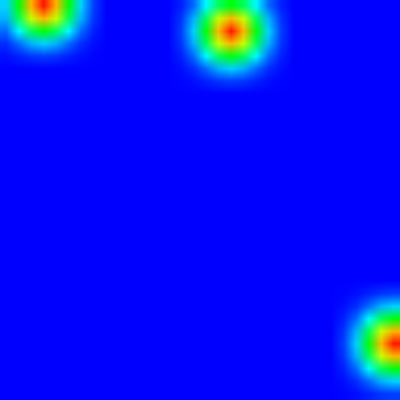}
} \hspace{0.2cm}
\subfloat[][\emph{Gaze heat-map}]{\label{fig:hm_synt_1}
    \includegraphics[height=0.27\linewidth]{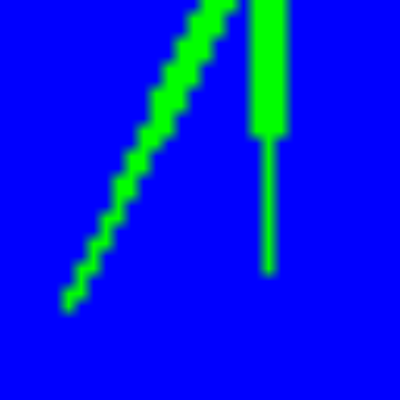}
} \hspace{0.2cm}
\subfloat[][Mean \emph{gaze heat-map}]{\label{fig:m_hm_synt_1} 
    \includegraphics[height=0.27\linewidth]{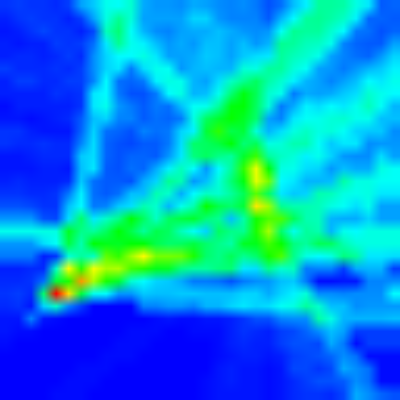}
}
\caption[Heat-maps from an example synthetic scenario ($N=2$ and $M=3$)]{Heat-maps from a synthetic scenario generated randomly, with $2$ people ($N=2$) and $3$ objects ($M=3$).~\subref{fig:o_synt_1}: the ground truth \emph{Object heat-map} $\Omegavect$ used for training or evaluation.~\subref{fig:hm_synt_1}: a \emph{Gaze heat-map} randomly chosen among the sequence.~\subref{fig:m_hm_synt_1}: the mean \emph{gaze heat-map} over the sequence.
}
\label{fig:illustration_synt_1}

\end{figure}


The Fig.~\ref{fig:illustration_synt_1} represents a synthetic scenario generated using this process. In practice, a wide variety of scenarios can be obtained with this approach. For instance, there is no limit to the number of people and/or objects that could be generated in one scenario, except the plausibility of such a scenario with respect to the physical space.

%% file: sec_experiments.tex
\section{Experiments}
\label{sec:gf-expe}

Experiments have been performed both on \synt data, generated online as described in section~\ref{sec:synthetic}, and on the \vern dataset~\cite{jayagopi2013vernissage} as described below. 
  Note that, we do not use the datasets employed in ~\cite{Cohen2012} and ~\cite{Brau_2018_ECCV} since they are not publicly available

\paragraph{The \vern Dataset}
\label{sec:vernissage-dataset}
It is composed of ten recordings lasting approximately ten minutes each. Each sequence contains two people interacting with a Nao robot and discussing about three wall paintings ($M=3$). The robot plays the role of an art guide, describing the paintings and asking questions to the people in front of it. 
The scene was recorded at 25 frames per second (fps) with an RGB camera embedded into the robot head, and with a VICON motion capture system consisting of a network of infrared cameras providing accurate position and head pose estimations of the people, the objects and the robot. We use the OpenCV version of \cite{Viola2001} for face detection and \cite{ba2016line} to track the faces over time. The head poses are estimated by employing \cite{lathuiliere17}. The 3D head positions, are estimated using the face center and the bounding-box size, which provides a rough estimate of the depth. The position of the robot itself and the orientation of its head are also known. Finally, the object locations are annotated along with the visual focus of attention of the participants over time. Images extracted from Nao camera during various recordings are displayed in Fig.~\subref*{fig:gf-VFOA},~\subref*{fig:gf-OFVOD},~\ref{fig:gf-outline},~\subref*{fig:camera10},~\subref*{fig:camera30},~\subref*{fig:camera80}.


\paragraph{Implementation details}

The heat-map dimensions are set to $S_X=S_Y=32$, to represent a room of size $3m \times 3m$. The cone aperture $\epsilon$ is set to $2\degree$. We fixed the input sequence size to $T=200$ time steps. On the \vern dataset, the videos are subsampled to 5 fps, then the duration of a sequence is $40s$ and we can extract several sequences from each video sequence. By using a sliding window and $50\%$ overlap, we extract a total of 224 sequences. We use the visual focus of attention annotations to obtain the true objects of interest for each sequence. Consequently, the number of objects can vary from 1 to 3 in the test sequences. We employ the adam optimizer \cite{kingma2014adam} for 10 epochs. For all neural network architectures employed in the experiments, the batch size is set to 32. In all cases, we perform the local maxima extraction method described in~\eqref{eq:NMS} after estimating $\hat{\Omegavect}$ to obtain the list of object positions. The neighborhood $\mathpzc{N}(\cdot)$ from~\eqref{eq:NMS} is defined as a sliding region of $5\times 5$ pixels, and the shrinking function $\alpha: x \mapsto \ln(1+x)$ . In all our experiments, we report \emph{Precision} and \emph{Recall}, and these two metrics are combined to obtain the \emph{f1-score}. \emph{Precision} measures the percentage of detected objects that are true objects. \emph{Recall} measures the percentage of true objects correctly detected. In order to compute these metrics, we employ a Hungarian algorithm that matches the detections with the real objects positions based on their respective distances. Importantly, the detection is considered as a success if the distance between the estimated and annotated distances is lower than $50$cm in the real-world space. For all learning-based approaches, we also report the MSE between the predicted and true \emph{object heat-maps}. Since Heuristic methods do not intend to predict the \emph{object heat-maps}, the MSE is not reported for them.

\paragraph{Results and Discussion}

\begin{table}

\caption[Gaze following performances on \synt data and \vern]{
  Results obtained on data from the proposed synthetic generator and on the \vern dataset \cite{jayagopi2013vernissage}. MSE values reported were multiplied by $10^{2}$ to facilitate reading. \emph{Precision}, \emph{recall} and \emph{f1-score} represent percentages. For learning-based approaches, we report the mean and standard deviation over five runs. 
    Results on the \cite{Brau_2018_ECCV} dataset are reported for comparison.
  \label{tab:results}
}

\begin{center}
  \resizebox{0.99\linewidth}{!} {
    \begin{tabular}{l|clll}

  	  \midrule
      Dataset & \multicolumn{4}{c}{Synthetic}\\

      \midrule
      Method & \emph{MSE} & \multicolumn{1}{c}{\emph{Precision}}& \multicolumn{1}{c}{\emph{Recall}} & \multicolumn{1}{c}{\emph{f1-score}}\\
      \midrule
      \midrule
      \emph{Cone}         & -  & 18.8 & 53.9 & 27.8 \\
      \emph{Intersect}    & -  & 21.1 & 35.0 & 26.3 \\

      \midrule
      \emph{Linear Reg.}  &    1.25 $\pm$ 0.02 &    50.5 $\pm$ 2.2 &    76.9 $\pm$ 1.0 &    60.9 $\pm$ 1.8\\
      \emph{1-FC}         &    1.06 $\pm$ 0.03 &    64.9 $\pm$ 1.6 &    61.5 $\pm$ 1.5 &    63.1 $\pm$ 1.1\\
      \emph{3-FC}         &    1.05 $\pm$ 0.01 &    65.9 $\pm$ 0.6 &    59.9 $\pm$ 2.2 &    62.8 $\pm$ 1.2\\

      \midrule
      \emph{Mean-2D-Enc}  &    1.00 $\pm$ 0.03 &    74.5 $\pm$ 2.4 &    59.5 $\pm$ 1.7 &    66.1 $\pm$ 1.3\\
      \emph{2D-Enc}       &    0.98 $\pm$ 0.02 &    76.8 $\pm$ 2.2 &    62.2 $\pm$ 1.5 &    68.7 $\pm$ 1.7\\
      \emph{3D-Enc}       &    0.85 $\pm$ 0.06 &    88.2 $\pm$ 3.9 &    71.4 $\pm$ 2.1 &    78.9 $\pm$ 2.4\\
      \emph{3D/2D U-Net}  &\bf 0.75 $\pm$ 0.01 &\bf 89.0 $\pm$ 1.2 &\bf 78.0 $\pm$ 0.6 &\bf 83.2 $\pm$ 0.8\\
      \midrule
      \midrule
      \midrule





      Dataset  &  \multicolumn{4}{c}{Vernissage}\\

      \midrule
      Method & \emph{MSE} & \multicolumn{1}{c}{\emph{Precision}}& \multicolumn{1}{c}{\emph{Recall}} & \multicolumn{1}{c}{\emph{f1-score}} \\
      \midrule \midrule
      \emph{Cone}         & -   & 20.7 & 35.8 & 26.2\\
      \emph{Intersect}    & -   & 34.9 & 27.2 & 30.6\\

      \midrule
      \emph{Linear Reg.}  &    1.48 $\pm$ 0.04 &    37.0 $\pm$ 4.9 &\bf 53.7 $\pm$ 5.0 &    43.7 $\pm$ 4.6\\
      \emph{1-FC}         &    1.49 $\pm$ 0.02 &    29.9 $\pm$ 3.2 &    35.2 $\pm$ 2.5 &    32.3 $\pm$ 2.8\\
      \emph{3-FC}         &    1.49 $\pm$ 0.02 &    28.0 $\pm$ 3.5 &    29.9 $\pm$ 1.5 &    28.8 $\pm$ 2.4\\

      \midrule
      \emph{Mean-2D-Enc}  &\bf 1.37 $\pm$ 0.02 &\bf 60.1 $\pm$ 1.5 &    41.1 $\pm$ 1.0 &\bf 48.8 $\pm$ 1.2\\
      \emph{2D-Enc}       &    1.39 $\pm$ 0.03 &    54.9 $\pm$ 4.2 &    40.5 $\pm$ 1.6 &    46.6 $\pm$ 2.5\\
      \emph{3D-Enc}       &    1.43 $\pm$ 0.05 &    49.9 $\pm$ 8.1 &    37.1 $\pm$ 9.0 &    42.5 $\pm$ 8.7\\
      \emph{3D/2D U-Net}  &    1.44 $\pm$ 0.04 &    45.1 $\pm$ 4.8 &    38.5 $\pm$ 2.2 &    41.5 $\pm$ 3.3\\

      \midrule
      \midrule
      \midrule
      Dataset  &  \multicolumn{4}{c}{Brau et al.~\cite{Brau_2018_ECCV}}\\
      \midrule
      Method & \emph{MSE} & \multicolumn{1}{c}{\emph{Precision}}& \multicolumn{1}{c}{\emph{Recall}} & \multicolumn{1}{c}{\emph{f1-score}} \\
      \midrule
      Brau et al.~\cite{Brau_2018_ECCV} & - & 59.0 & 48.0 & \it 52.9\\

      \bottomrule
    \end{tabular}
  }

\end{center}

\end{table}


In Table \ref{tab:results}, we report the results obtained employing all methods described on both \synt and real data. 

It has to be noted that many different recurrent architectures have been considered, either alone or in conjunction with one of the proposed convolutional Encoder/Decoder architectures \eg adapted from the convolutional LSTM~\cite{donahue2015long}. All of them converged to networks predicting always the same (or almost the same) \emph{object heat-map}. We believe that, in this formulation, the ability to combine information from distant time frame is important, and this is difficult to achieve with RNN (or LSTM) processing data sequentially \cite{pascanu2013difficulty}. 

\begin{figure}
\centering     
\subfloat[][$\hat{\Omegavect}$ - \emph{Mean-2D-Enc}]{\label{fig:omega_2D}
    \includegraphics[height=0.30\linewidth]{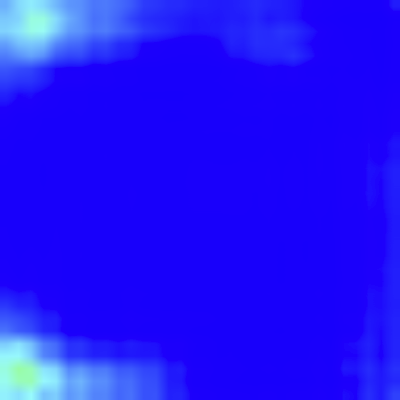}
}
\subfloat[][$\hat{\Omegavect}$ - \emph{3D/2D U-Net}]{\label{fig:omega_3D}
    \includegraphics[height=0.30\linewidth]{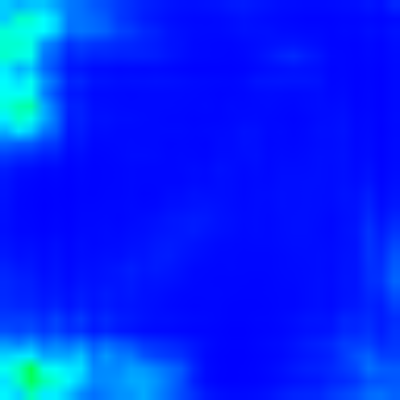}
}
\subfloat[][$\hat{\Omegavect}$ - \emph{Linear Reg.}]{\label{fig:omega_lin}
    \includegraphics[height=0.30\linewidth]{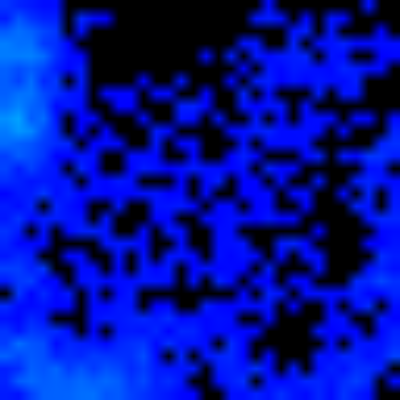}
}\\
\subfloat[][Obj - \emph{Mean-2D-Enc}]{\label{fig:obj_2D}
    \includegraphics[height=0.30\linewidth]{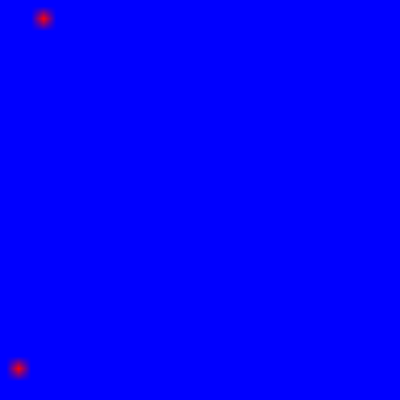}
}
\subfloat[][Obj - \emph{3D/2D U-Net}]{\label{fig:obj_3D}
    \includegraphics[height=0.30\linewidth]{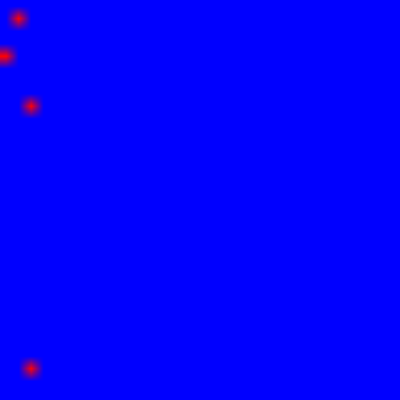}
}
\subfloat[][Obj - \emph{Linear Reg.}]{\label{fig:obj_lin}
    \includegraphics[height=0.30\linewidth]{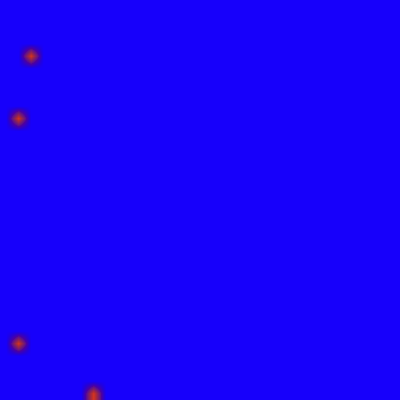}
}

\caption[]{Results of three methods on the \vern scenario illustrated in Fig.~\ref{fig:illustration}.
  \subref*{fig:omega_2D}, \subref*{fig:omega_3D}, \subref*{fig:omega_lin}: Estimates $\hat{\Omegavect}$ of the \vern \emph{object heat-map} $\Omegavect$ from Fig.~\subref*{fig:o_hm} using three different architectures.
  \subref*{fig:obj_2D}, \subref*{fig:obj_3D}, \subref*{fig:obj_lin}
  : Corresponding objects positions, obtained as the highest local maxima from $\hat{\Omegavect}$.
  Black pixels in~\subref*{fig:omega_lin} 
  indicate negative values.
}
\label{fig:results}


\end{figure}


From the experiments, we observe that learning-based approaches clearly outperform those based on cone intersections inspired from \cite{Cohen2012}. Indeed, even on the \synt datasets, their \emph{precision} and \emph{recall} do not reach better than $18.8\%$ and $53.9\%$ respectively, whereas a simple linear regression reaches considerably higher scores ($50.5\%$ and $76.9\%$ respectively). The same remark stands for the \vern dataset. Increasing the network complexity by simply adding fully-connected layers does not bring any improvement and even reduce the performance. Then, we observe that all proposed encoder/decoder models clearly outperform other methods by a substantial margin on the \synt dataset. There, we obtain a $22.3\%$ gain in terms of f1-score when employing the \emph{3D/2D U-Net} with respect to the linear regression model. On the \vern dataset, a $5.1\%$ gain is obtained in terms of \emph{f1-score} when employing the \emph{Mean-2D-Enc} with respect to the linear regression model. These experiments validate the use of the encoder/decoder architecture.

We notice that the performance on the \synt dataset increases with encoder complexity. However, the inverse phenomenon is observed on \emph{Vernissage}, where the best performances are obtained using the simplest encoder architecture (that does not model time). 
Our guess for this observation is that there is a significant discrepancy between the distribution of \vern data and the \synt data distribution sampled according to \eqref{eq:sample}. 
Therefore, more complex models probably tend to over-fit the \synt data distribution, and thus transfer less well on the \vern dataset. 
More realistic training data could lead to further improvements. This could be obtained by gathering a dataset of real-life scenarios which could be use either as training data or to improve the quality of the generative model.

The only methods from the literature that we are aware of are~\cite{Cohen2012} and~\cite{Brau_2018_ECCV}. In both cases, neither the data nor the code have been made available online. Moreover, the papers lack information about parameters or hyperparameters that prevented us to test it. Additionally, \cite{Brau_2018_ECCV} explicitly discarded the \vern dataset in their experiments. Results on their dataset (59\% precision and 48\% recall) are comparable to ours on \emph{Vernissage}. Note that, \cite{Brau_2018_ECCV} employed a larger success threshold ($1.0$m in
the real-world space for $50$cm in our case) and consequently would obtain lower scores according to our evaluation protocol. We wish to test our method on their dataset in the future. We do not compare to~\cite{Cohen2012} since they did not report any quantitative results on location estimation.

In Fig.~\ref{fig:results}, the predicted \emph{gaze heat-maps} $\hat{\Omegavect}$ for several learning-based approaches applied on the scenario from Fig.~\ref{fig:illustration} are displayed. The architectures \emph{Mean-2D-Enc} and \emph{Linear Reg.} use the average \emph{gaze heat-map}  $\frac{1}{T} \sum_{t=1}^T \Gammavect_t$ as input, whereas \emph{3D/2D U-Net} takes the whole concatenated sequence $\Gammavect_{1:T}$. All three approaches are approximately able to predict the positions of two objects of interest. The third object is probably not targeted enough during the sequence to be found. The black pixels in the \emph{Linear Regression} indicate negative values. All other approaches end with a sigmoid activation so each pixel value is homogeneous to a probability. The lower number of falsely proposed object positions for the \emph{Mean-2D-Enc} is consistent with the higher mean precision reported.
For comparison, experiments on the \synt scenario from Fig.~\ref{fig:illustration_synt_1} are available in the supplementary materials.

%% file: sec_conclusion.tex
\section{Conclusions}
\label{sec:gf-concl}


In this paper, we define the problem of \emph{extended gaze following} as finding the locations of objects of interest solely from the gaze direction of visible people. Importantly, this allows for finding objects either inside or outside the camera field-of-view. In this context, we propose a novel spatial representation for head poses (approximating gaze direction) and object locations. 
We present a framework that takes advantage of convolutional encoder/decoder architectures to learn the spatial relationship between head poses and object locations, and we compare nine different methods on synthetic and real data. We finally conclude that learning-based approaches outperform geometry-based ones while being competitive with the state of the art. We also show that the necessary training examples can be quickly and easily obtained through a synthetic data generation process.

We believe this work open new perspectives for research. In particular, several decisions were taken to obtain an end-to-end method (\eg heat-map representation or elevation coordinate omission), which makes it hardly suitable in some situations. The extended gaze-following problem would benefit greatly from a benchmark of different representations and inference models, and of the influence of each simplifying hypothesis. Moreover, the availability of suitable datasets would ease future research on this topic.
In parallel, we wish to use this framework in the future as a tool to improve the decision process of a robotic system in a social context such as~\cite{lathuiliere2019prl}.